\documentclass[Afour,sageh,times]{sagej}

\usepackage{moreverb,url}

\usepackage[colorlinks,bookmarksopen,bookmarksnumbered,citecolor=red,urlcolor=red]{hyperref}

\newcommand\BibTeX{{\rmfamily B\kern-.05em \textsc{i\kern-.025em
b}\kern-.08em T\kern-.1667em\lower.7ex\hbox{E}\kern-.125emX}}

\def\volumeyear{2022}

\usepackage{float}
\usepackage{natbib} \usepackage{mathtools}

\DeclarePairedDelimiter\floor{\lfloor}{\rfloor}

\usepackage[nolist]{acronym} \usepackage{etoolbox} \makeatletter
\newif\if@in@acrolist \AtBeginEnvironment{acronym}{\@in@acrolisttrue}
\newrobustcmd{\LU}[2]{\if@in@acrolist#1\else#2\fi}
\newcommand{\ACF}[1]{{\@in@acrolisttrue\acf{#1}}} \makeatother 

\newcommand{\bm}[1]{\boldsymbol{#1}}
\renewcommand{\bf}[1]{\textbf{#1}}

\begin{document} \runninghead{Daryanavard and Porr}

\title{Sign and relevance learning}

\author{Sama Daryanavard\affilnum{1} and Bernd Porr\affilnum{1}}

\affiliation{\affilnum{1}Biomedical Engineering Division, School of
Engineering, University of Glasgow, Glasgow G12 8QQ, UK.}

\corrauth{Sama Daryanavard, University of Glasgow, Glasgow G12 8QQ,
UK.}

\email{sama.daryanavard@glasgow.ac.uk}
\begin{acronym}
\setlength{\parskip}{0ex} \setlength{\itemsep}{0ex}
\acro{GSV}[GSV]{\LU{G}{g}rey-\LU{S}{s}cale \LU{V}{v}alue}
\acro{RPi}[RPi]{Raspberry Pi 3B+} \acro{cE}[$E$]{\LU{C}{c}ontrol
\LU{E}{e}rror} \acro{LDR}[LDR]{\LU{L}{l}ight \LU{D}{d}ependent
\LU{R}{r}esistor} \acro{MC}[MC]{\LU{M}{m}otor \LU{C}{c}ommand}
\acro{FIR}[FIR]{\LU{F}{f}inite \LU{I}{i}mpulse \LU{R}{r}esponse}
\acro{FA}[FA]{\LU{F}{f}ilter \LU{A}{a}rray}
\acro{FoV}[FoV]{\LU{F}{f}ield of \LU{V}{v}iew}
\acro{G}[G]{\LU{G}{g}rey \LU{V}{v}alue} \acro{SaR}[SaR]{\LU{S}{s}ign
and \LU{R}{r}elevance} \acro{ltp}[LTP]{\LU{L}{l}ong-\LU{T}{t}erm
\LU{P}{p}otentiation} \acro{ltd}[LTD]{\LU{L}{l}ong-\LU{T}{t}erm
\LU{D}{d}epression} \acro{GDM}[GDM]{\LU{G}{g}radient \LU{D}{d}escent
\LU{M}{m}ethod}
\end{acronym}

\begin{abstract}

Standard models of biologically realistic or biologically inspired
reinforcement learning employ a global error signal, which implies the
use of shallow networks. On the other hand, error backpropagation
allows the use of networks with multiple layers. However, precise
error backpropagation is difficult to justify in biologically
realistic networks because it requires precise weighted error
backpropagation from layer to layer. In this study, we introduce a
novel network that solves this problem by propagating only the sign of
the plasticity change (i.e., LTP/LTD) throughout the whole network,
while neuromodulation controls the learning rate. Neuromodulation can
be understood as a rectified error or relevance signal, while the
top-down sign of the error signal determines whether long-term
potentiation or long-term depression will occur. To demonstrate the
effectiveness of this approach, we conducted a real robotic task as
proof of concept. Our results show that this paradigm can successfully
perform complex tasks using a biologically plausible learning
mechanism.

\end{abstract}


\maketitle

\section{Introduction\label{intro}}

The learning of an organism can be understood in the context of its
interactions with the environment, facilitated through sensory inputs
and motor outputs, which, in turn, lead to new sensory inputs
\citep{maffei2017perceptual}. The framework for such learning is based
on closed-loop learning \citep{Uexkuell26}, where actions result in
either positive or negative consequences. This is the realm of
reinforcement learning \citep{Dayan2002Neuron}. The reward prediction
error is central to reinforcement learning in a biologically realistic
framework. In the 1990s, \citet{Schultz97} suggested that dopamine
codes this error \citep{Bromberg2010,wood2017,Takahashi2017}, which is
similar to the temporal difference error in machine learning
\citep{Sutton88}. This led to the assumption that the brain resembles
an actor/critic architecture, where dopamine, as the reward prediction
error, drives synaptic changes in the striatum \citep{Humphries2006}.
Another interpretation of the actor/critic architecture is that it
represents a nested closed-loop platform, where an inner reflex loop
generates an error signal that tunes an actor in an outer loop to
create anticipatory actions. In other words, the actor generates a
forward model of the reflex \citep{porr2002isotropic}. Therefore, the
actor/critic architecture can be used for both model-based
\citep{verschure1991adaptive} and model-free learning.

However, relying solely on a global error signal, such as the dopamine
reward prediction error, has its limitations as this error signal
affects all neurons \citep{Humphries2006,oreilly2006}, thus making
multi-layer networks less useful. For this reason, while the striatum
may be a single-layer structure that receives dopamine from the
substantia nigra pars compacta (SNc), cortical networks, such as the
orbitofrontal (OFC) and medial prefrontal cortices (mPFC), are also
heavily involved in reinforcement learning and decision making
\citep{haber95,berthoud04,Rolls2008,DelaCruz2016}, being innervated by
dopaminergic neurons from the ventral tegmental area (VTA).

In the context of cortical pyramidal neurons, it is important to
remember that the main method by which plasticity is induced is
through learning rules such as Hebbian learning \citep{Hebb49,Bliss73}
or spike-timing-dependent plasticity (STDP) \citep{Markram97}, with
its underlying postsynaptic calcium dynamics where a high
concentration causes \ac{ltp} and a low concentration causes \ac{ltd}
\citep{Lu2001,Castellani01}. Therefore, calcium determines the
\textsl{sign} of the plasticity, whether it is \ac{ltp} or \ac{ltd}.
These learning rules interact with the global neuromodulator
\citep{Mattson2004,Lovinger2010}, particularly serotonin
\citep{Roberts.2011,Linley2013,Luo2015,Li2016,Crockett2009}, but also
dopamine \citep{DelaCruz2016}, which can be thought of as controlling
the overall \textsl{learning rate}. From a theoretical point of view,
this has inspired ISO3 learning, where differential Hebb is combined
with a rectified error signal called ``relevance''. Nonetheless, this
network is also shallow, consisting of just one layer.

This paper presents a novel learning mechanism that expands upon the
single-layer approach of ISO3 learning \citep{Porr2007} by
incorporating multiple layers. Our network utilises a top-down pass of
the \textit{sign} of an error signal to determine whether to implement
\ac{ltp} or \ac{ltd}, while a global neuromodulator controls the
learning speed. As a proof of concept, we apply our mechanism to a
simple line-following task on a real robot. The robot's deviation
generates an error signal that is used to train the network.

\section{The \acf{SaR} learning platform}

\begin{figure}[t]
\centering
\includegraphics[width=0.5\textwidth]{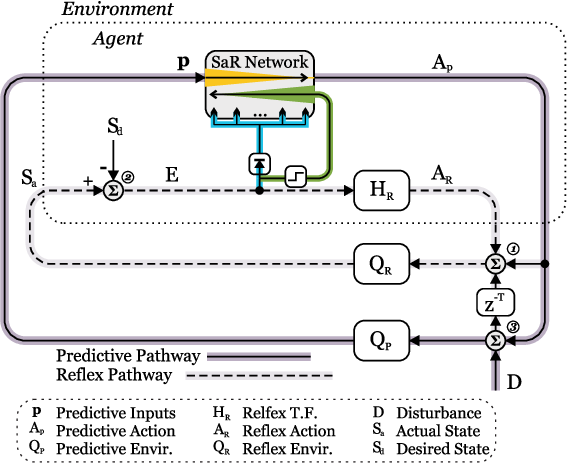}
\caption{\textit{The \ac{SaR} learning platform. It is comprised of
two key components, the agent and the environment, which are
demarcated by a dotted rectangle representing their boundary. The
platform consists of an inner reflex loop, represented by dashed
lines, and an outer predictive (or learning) loop, represented by
solid lines.}}\label{fig:fig1SAR_platform}
\end{figure}

The learning platform for \acf{SaR} is depicted in
Figure~\ref{fig:fig1SAR_platform}. The underlying concept involves a
fixed feedback controller, which endeavours to maintain a desired
state and generates an error signal by comparing its desired state
with the actual state. For instance, this may involve aiming for a
target location. The error signal is then used to facilitate learning
in another adaptive loop \citep{Verschure91,PorrNecoISO2003}.
Essentially, this adaptive loop learns the forward model of the fixed
control loop, which is referred to as the ``reflex'' mechanism. The
transfer functions $Q_{R}$ and $Q_{P}$ correspond to the environment
for reflex and predictive mechanisms, respectively. Similarly, $H_{R}$
corresponds to the transfer function of the reflex mechanism itself.
These functions translate actions into sensory consequences ($Q_{R}$
and $Q_{P}$), and vice versa ($H_{R}$). In the context of control
systems, these functions are measured using system identification
techniques, for example measured input-output data \citep{Chen2017}.
In the context of model-based learning, however, the network readily
generates the forward model of the environment. The system does not
rely on any previous knowledge of the environment and thus these
functions are not computed in this work. The reflex mechanism, the
learning pathway, and the flow of signals are described
in detail below.

\subsection{The reflex loop}

Reflexes are among the most innate drivers of an organism's behaviour.
They are triggered involuntary actions that are carried out without
prior knowledge of the stimuli. Typically, a reflex serves to ensure
the survival and success of the agent. In this work, the agent is a
navigational robot that aims to follow a path. When encountering a
bend in the path, this innate reflex mechanism provides the organism
with an immediate reaction that is designed to recover from the
disturbance. Therefore, a reflex can be thought of as a fixed
closed-loop controller that opposes collective disturbances while
maintaining its desired state.

The desired state, denoted as $S_d$ located at node
\textcircled{\small{2}}, represents the state that the agent aims to
achieve in order to maximise its reward. In error-based learning,
which is the method employed in this study, $S_d$ corresponds to the
state where the punishment is minimised instead, meaning that the
error is zero. Typically, the desired state is specified as a goal or
objective for the agent to pursue. When the agent reaches this state,
it achieves inner equilibrium and stability despite any external
disturbances, and therefore no further action is required from the
agent.

When a disturbance $D$ propagates through the reflex environment $Q_R$
(refer to Figure~\ref{fig:fig1SAR_platform}), it triggers a change in
the agent's state. This actual state of the agent, denoted as $S_a$,
is continuously compared to the desired state, and any discrepancies
between the two are detected by the agent's sensors. Such
discrepancies generate an error signal known as the \acf{cE}:

\begin{align}\label{eq:cError}
\ac{cE} = S_{d} - S_{a}
\end{align}

In technical terms, the desired state is the state in which the error
signal is zero. However, if the error signal is non-zero, the reflex
mechanism $H_{R}$ will take an action $A_{R}$ designed to correct for
the disturbance and return the agent to equilibrium at node
\textcircled{\small{1}}. This summarises the reflex pathway. Although
this loop is designed to combat disturbances, it is incapable of
avoiding them altogether. To maintain the desired state at all times,
the reflex loop is enclosed within an outer predictive loop whose task
is to mitigate the effects of the disturbance $D$ before it reaches
the reflex loop.

\subsection{The predictive loop}

In contrast to the reflex, the \acf{SaR} network anticipates and
accounts for the disturbance with prior knowledge. This is made
possible by structuring the environment so that the disturbance $D$ is
delayed by $z^T$ time-steps, allowing the learner to generate
anticipatory actions. The learner's goal is to interact with the
environment and achieve the objective without invoking the reflex
mechanism. In this work, the learner aims to follow the path without
relying on the reflex sensors.

\begin{figure}[H]
\centering
\includegraphics[width=0.45\textwidth]{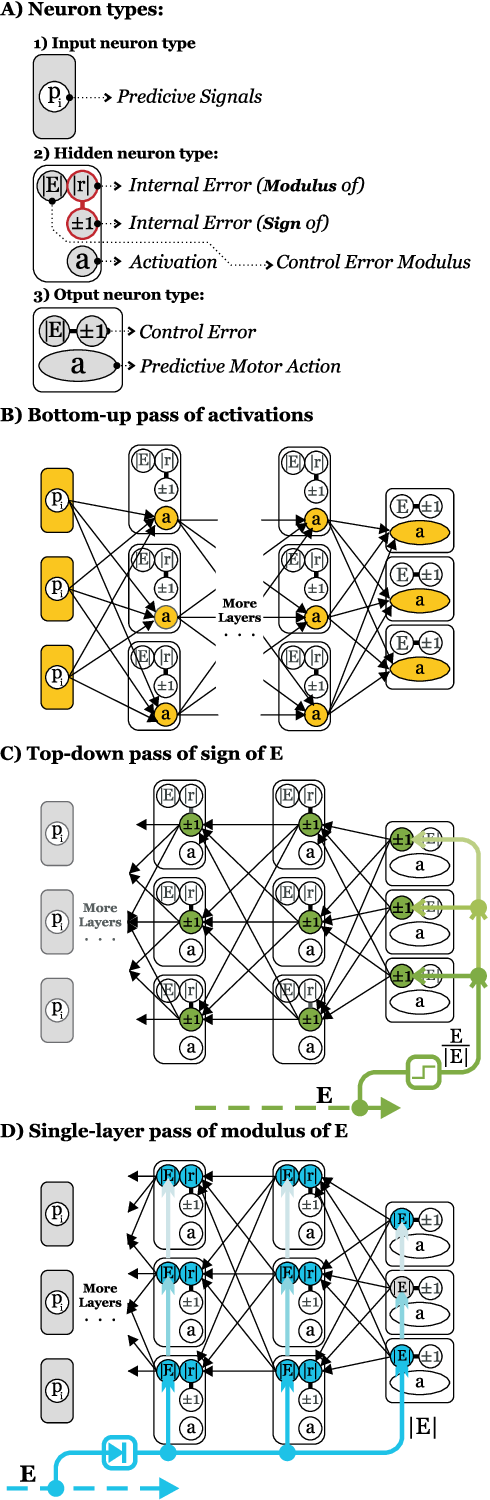}
\caption{\textit{Signal pathways within the \ac{SaR} network: A)
Symbol introduction, including an input neuron, a hidden neuron, and
an output neuron. B) The bottom-up pass of predictive signals, with
calculations of activations and formation of motor commands. C) The
top-down pass of the sign of \ac{cE}. D) The single-layer pass of
\ac{cE}.}}\label{fig:fig2SAR_threepass}
\end{figure}

The disturbance is first received by the learner's environment $Q_{P}$
and translated into predictive sensory inputs $\bf{p}$. Based on this
input information, the network generates a predictive action $A_{P}$
in an attempt to restore system equilibrium at node
\textcircled{\small{1}}, where the effects of $A_{P}$, the delayed
$D$, and the reflex action $A_{R}$ are combined before travelling
through to the reflex loop. If successful, $A_{P}$ counters the
delayed $D$ precisely to yield zero at this summation node, and thus
the reflex loop is not evoked\footnote{Note that $A_{R}$ is zero at
this instance.}. However, if unsuccessful, the summation yields a
non-zero signal that is received by $Q_{R}$, evoking the reflex loop
and resulting in a non-zero \acl{cE} signal.

\subsection{The \acf{cE}}

The function of \acl{cE} is two-fold: 1) it is received by the reflex
$H_{R}$ to generate the reflex action $A_{R}$, as described above, and
2) it serves as instructive feedback for the learner. Through
iterations, the non-zero \ac{cE} signal tunes the internal parameters
of the \ac{SaR} network. The learning terminates when $A_{P}$
precisely and persistently counters the disturbance at node
\textcircled{\small{3}}. In this case, the reflex loop is no longer
evoked, \ac{cE} remains at zero, and the learner undergoes no further
changes. Thus, the \ac{SaR} network will have successfully generated
the forward model of the reflex, similar to the model-based learning
paradigms presented in the work by
\citet{porr2002isotropic,porr2006strongly,porr2003isotropic}.

\begin{figure*}[t]
\centering
\includegraphics[width=0.75\textwidth]{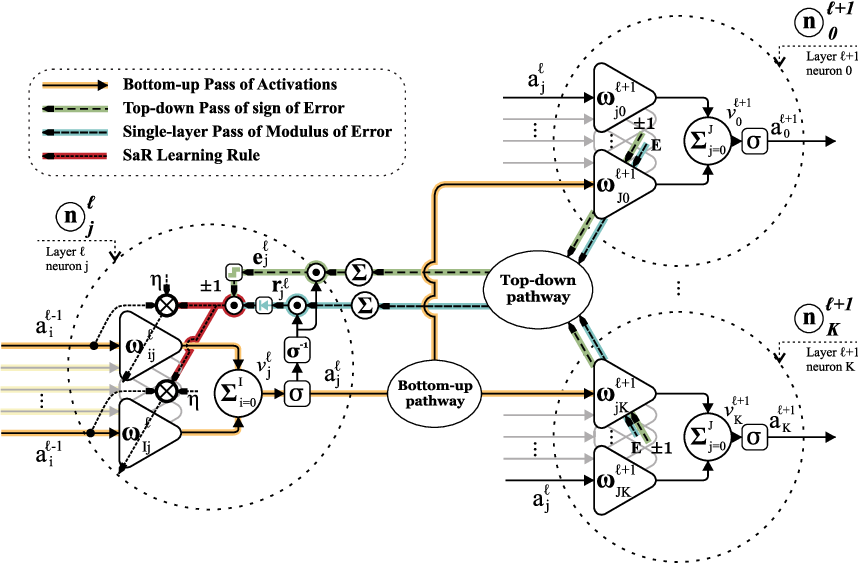}
\caption{\textit{Inner connections of neurons in a \acf{SaR} network.
Yellow arrows indicate the forward pass of the activations, green
arrows represent the top-down pass of the sign of error, blue arrows
denote the single-layer pass of the modulus of error, and red arrows
highlight the learning process.}}\label{fig:fig3SAR_neuron}
\end{figure*}

\subsection{The \ac{SaR} learner}

The learner generates the forward model of the reflex by mapping its
predictive actions onto a set of sensory consequences, which is
realised by the agent - the \acl{cE} signal. The unique property of
the \ac{SaR} paradigm is that the instructive feedback of \ac{cE} is
facilitated through two distinct pathways: 1) the top-down pass of the
\textit{sign} of \ac{cE} (green traces), and 2) global intervention of
the rectified \ac{cE} (blue traces); see
Figure~\ref{fig:fig1SAR_platform}.

Figure~\ref{fig:fig2SAR_threepass} illustrates these pathways in
detail. The symbols used for each neuron type are introduced in
Panel~\ref{fig:fig2SAR_threepass}A, which includes neurons in the
input, hidden, and output layers. The input neurons receive the
predictive signal to be injected into the network. The hidden layers
contain the \acl{cE}, their internal error (sign and modulus of), and
their activation. Finally, the output neurons contain the \acl{cE} and
the final predictive motor action of the network.

The \ac{SaR} learner employs a conventional feed-forward neural
network with fully-connected layers.
Panel~\ref{fig:fig2SAR_threepass}B shows the forward pass of signals
in the network, from the predictive inputs $P_{i}$ to activations $A$
in hidden layers and predictive outputs $A_{pi}$ in the output layer.

Panel~\ref{fig:fig2SAR_threepass}C shows the top-down pass of the sign
of the error \ac{cE}. The green traces mark the entry of the error
\ac{cE} from the reflex loop onto the output neurons. The sign of the
error \ac{cE} is passed from the final layer to the deeper layers.
Within each layer, the sign of the resulting value is passed to the
deeper layers. This results in an error of $\pm 1$ within each neuron,
which primes their connections to be strengthened ($+1$) or weakened
($-1$). This is analogous to \ac{ltp} and \ac{ltd} in the context of
neurophysiology.

Figure~\ref{fig:fig2SAR_threepass}D illustrates the single-layer
traversal of the modulus of \ac{cE}. The blue traces represent the
entry of this signal from the closed-loop platform into each layer.
This value activates each neuron and is transmitted only to its
adjacent deeper neurons. The absolute value of the resulting sum
determines the extent to which the previously primed connections are
strengthened or weakened. This mechanism is similar to the effect of
neuromodulators on plasticity, particularly serotonin.

\section{Mathematical derivation of \acf{SaR} learning}

In this section, we derive the learning rule for the \acf{SaR}
paradigm. First, the bottom-up pass of predictive inputs is derived,
following the conventional flow of signals in fully-connected
feed-forward neural networks. Next, a mathematical expression of the
learning goal in its general sense is presented, where a
differentiable function is optimised through adjustments of weights.
Subsequently, this learning goal is unravelled with respect to the
closed-loop platform and the inner workings of the neural network.
This leads to the derivation of the learning rule for a conventional
\acf{GDM}, which in turn provides the formulation of the \ac{SaR}
update rule.

\subsection{Bottom-up pass of activations}

Figure~\ref{fig:fig3SAR_neuron} illustrates the inner components of
neurons and their connections. The bottom-up pass of activations, from
the $j^{th}$ neuron in the $\ell^{th}$ layer,
$\textcircled{\small{$n$}}_{j}^{\ell}$, to all neurons in the adjacent
deeper layer, $\textcircled{\small{$n$}}_{0 \rightarrow K}^{\ell+1}$,
is highlighted by yellow solid lines. The figure shows the network's
parameters in scalar mode, while for mathematical derivation, a matrix
representation of parameters is adopted. Equations~\ref{eq:frwdProp1}
and~\ref{eq:frwdProp2} summarise the forward pass from the input
layer, through hidden layers, and to the output layer:

\begin{align}\label{eq:frwdProp1}
\bf{a}^{1} &= \sigma(\bf{v}^{1}) = \sigma(\bm{\omega}^{1}
\cdot \bf{p}) &&\text{ Input layer}
\end{align}

\begin{align}\label{eq:frwdProp2}
\bf{a}^{\ell} &= \sigma(\bf{v}^{\ell}) =
\sigma(\bm{\omega}^{\ell} \cdot \bf{a}^{\ell-1}) &&\text{ Hidden layers }
\end{align}

The notation $\bf{p}$ represents the matrix of predictive
inputs to the network. Meanwhile, $\bf{a}^{\ell}$, $\bf{v}^{\ell}$, and
$\bm{\omega}^{\ell}$ denote the activation, sum-output, and weight
matrices within the $\ell^{th}$ layer, respectively. Depending on the
specific application, the predictive action may take the form of a
linear function of output activations:

\begin{align}
A_{P} = f(\bf{a}^{L})
\end{align}

As previously mentioned, when this action is executed, a sequence of
events occurs within the closed-loop platform that ultimately results
in the generation of the \ac{cE} signal, as defined in
Equation~\ref{eq:cError}. This signal plays a crucial role in
governing the learning process of the \ac{SaR} network.

\subsection{The learning goal}

The primary objective of the learning process is to actively and
consistently maintain the \acl{cE} at zero, which is accomplished by
adjusting the weight matrices. From a
mathematical perspective, this learning goal is most effectively
formulated as an optimisation task, in which the quadratic expression
of the \acl{cE} is minimised with respect to the weights.

\begin{align}\label{eq:goal}
\bm{\omega}^{1 \rightarrow L} =
\underset{\bm{\omega}}{\text{argmin}}~E^{2}
\end{align}

To put it simply, our goal is to find weight matrices for all layers
of the network such that the resulting action minimises $E^{2}$, which
is equivalent to driving $E$ towards zero. When framed as an
optimisation task, a commonly used technique is the \acf{GDM}, which
involves adjusting an arbitrary weight in proportion to the
sensitivity of $E^{2}$ with respect to the sum-output of the neuron
associated with that weight.

\begin{align}\label{eq:gdm}
\Delta \bm{\omega}^{\ell} = \frac{\partial
E^{2}}{\partial \bf{v}^{\ell}} \cdot \frac{\partial \bf{v}^{\ell}}{\partial
\bm{\omega}^{\ell}}
\end{align}

Referring to Equation~\ref{eq:frwdProp2}, the latter gradient yields
the matrix of activation inputs to the neuron, denoted as $\bf{a}^{\ell
-1}$. On the other hand, the former gradient establishes a
relationship between the closed-loop signal ($E$) and an internal
parameter of the network ($v$). The predictive action $A_{P}$
\citep{daryanavard2020closed} serves as the link between the
closed-loop platform and the neural network. Through the application
of the chain rule, the closed-loop signals are effectively separated
from the internal parameters of the network.

\begin{align}\label{eq:chainEPw}
\frac{\partial E^{2}}{\partial
\bf{v}^{\ell}} = \frac{\partial E^{2}}{\partial A_{P}} \cdot
\frac{\partial A_{P}}{\partial \bf{v}^{\ell}}
\end{align}

\subsection{Dynamics of the closed-loop platform}

Our objective is to derive an expression that establishes the
relationship between the \acf{cE} and the predictive output $A_{p}$.
By substituting for $S_{a}$ in Equation~\ref{eq:cError}, we obtain:

\begin{align}\label{eq:E}
E &= S_{d} - Q_{R}\overbrace{(A_{R} + A_{P}
+ Dz^{-T})}^{\text{Figure~\ref{fig:fig1SAR_platform}, node}
\textcircled{\small{1}}} \nonumber \\ &= S_{d} - Q_{R}(E H_{R} + A_{P}
+ Dz^{-T}) \nonumber \\ &= \frac{S_{d} - Q_{R}(A_{P} + Dz^{-T})}{1 +
Q_{R}H_{R}}
\end{align}

Hence, we arrive at:

\begin{align}\label{eq:dEdP}
\kappa := \frac{\partial E^{2}}{\partial
A_{P}} = 2 E \frac{\partial E}{\partial A_{P}} = 2 E \frac{-Q_{R}}{1 +
Q_{R}H_{R}}
\end{align} where $\frac{-Q_{R}}{1 + Q_{R}H_{R}}$ denotes
the reflex loop gain and is determined experimentally. This partial
gradient is known as the closed-loop gradient and is denoted by
$\kappa$.

\subsection{Inner dynamics of the neural network}

We aim to derive an expression that relates the predictive output
$A_{P}$ to the matrix of sum-outputs $\bf{v}^{\ell}$ in an arbitrary
layer. To this end, we can rewrite Equation~\ref{eq:frwdProp2} as
follows:

\begin{align}\label{eq:forPassRE}
\sigma(\bf{v}^{\ell + 1}) =
\sigma(\bm{\omega}^{\ell + 1} \cdot \sigma(\bf{v}^{\ell}))
\end{align}

Taking the derivative of $A_{P}$ with respect to $\bf{v}^{\ell}$ yields:

\begin{align}\label{eq:backprop}
\underset{\text{GDM}}{\bf{e}^{\ell}} :=
\frac{\partial A_{P}}{\partial \bf{v}^{\ell}} = \sigma^{-1}(\bf{v}^{\ell})
\cdot ({\bm{\omega}^{\ell+1}}^{T} \cdot \frac{\partial A_{P}}{\partial
\bf{v}^{\ell+1}})
\end{align}

The partial gradient resulting from this differentiation is referred
to as the internal error, and is denoted by the matrix
$\bf{e}^{\ell}$. It is important to note the recursive nature of
this operation, where the calculation of the partial gradient in the
$\ell^{th}$ layer depends on its calculation in the $(\ell + 1)^{th}$
layer. Therefore, starting from the final layer, we define the
function $f$ as the one that generates the predictive output given the
activation matrix of the output layer:

\begin{align}\label{eq:ap}
A_{P} = f(\sigma(\bf{v}^{L}))
\end{align}

Differentiation with respect to the sum-output matrix in the final
layer $v^{L}$ yields:

\begin{align}\label{eq:backprop3}
\bf{e}^{L} = \frac{\partial
A_{P}}{\partial \bf{v}^{L}} = f^{-1}(\sigma(\bf{v}^{L})) \cdot
\sigma^{-1}(\bf{v}^{L})
\end{align}

The computation of the internal error in the final layer triggers the
backwards top-down pass, which, in turn, yields the internal errors
for all layers. The learning rule for \ac{GDM} can be expressed as
follows:

\begin{align}\label{eq:gdmFULL}
\underset{\text{GDM}}{\Delta \bm{\omega}^{\ell}}
= \kappa \cdot \eta \cdot ( \overbrace{\bf{e}^{\ell}}^{\text{Internal
Error}} \cdot \bf{a}^{\ell - 1})
\end{align}

where $\eta$ is a scalar learning rate. The learning rule for \ac{SaR}
is derived in the following section.

\subsection{The \acf{SaR} learning rule}

In this work, we use only the \textit{sign} of the internal error
$\bf{e}^{\ell}$ for the top-down pass into deeper layers, as shown
by the green dashed arrows in Figure~\ref{fig:fig3SAR_neuron}. The
result is a matrix of signs\footnote{This matrix only contains values
of $+1, -1,$ or $0$.}, represented as $\bf{s}^{\ell}$ for each layer,
and calculated as follows:

\begin{eqnarray}
{\bf{e}}^{\ell} & = &\sigma^{-1}(\bf{v}^{\ell}) \cdot
({\bm{\omega}^{\ell+1}}^{T} \cdot \bf{s}^{\ell + 1})
\label{eq:diffusion} \\ {\bf{s}}^{\ell} & =
&\frac{\bf{e}^{\ell}}{|\bf{e}^{\ell}|} \label{eq:signcalc}
\end{eqnarray}

This sign matrix is used to prime the weights for updating. A sign of
$+1$ primes the corresponding weight to increase, while $-1$ causes
the associated weight to decrease. A sign of $0$ leaves the weight
unchanged. It is worth noting that the derivative of the sigmoid
function is strictly positive and, therefore, does not influence the
resulting sign matrix in Equation~\ref{eq:signcalc}. This
propagation spans the entire network, with only the resulting sign
transmitted through the layers.

Once the neurons are primed, the magnitude of their excitation or
depression is obtained from a single-layer pass of \acf{cE}. This is
illustrated by the blue dashed arrows in
Figure~\ref{fig:fig3SAR_neuron} and is formulated as follows:

\begin{align}\label{eq:gamma}
\bf{r}^{\ell} =
\sigma^{-1}(\bf{v}^{\ell}) \cdot ({\bm{\omega}^{\ell+1}}^{T} \cdot E)
\end{align}

where $\bf{r}^{\ell}$ is the matrix of relevance signal in the
$\ell^{th}$ layer. Unlike the internal error, the relevance signal is
not passed from, or to, deeper layers; the control error interrupts
this propagation. This means that $\bf{r}^{\ell}$ does not depend on
$\bf{r}^{\ell + 1}$. Therefore, this signal can be generated
simultaneously across all layers globally. For the final layer, it is
computed as:

\begin{align}\label{eq:singleLayerpropFINAL}
\bf{r}^{L} =
f^{-1}(\sigma(\bf{v}^{L})) \cdot \sigma^{-1}(\bf{v}^{L}) \cdot E
\end{align}

With this, the learning rule for the \ac{SaR} network is defined as:

\begin{align}\label{eq:sarFULL}
\underset{\text{\ac{SaR}}}{\Delta \bm{\omega}^{\ell}} = \kappa \cdot \eta
\cdot ((\overbrace{\overbrace{\bf{s}^{\ell}}^{\text{Sign}} \cdot
\overbrace{|\bf{r}^{\ell}|}^{\text{Relevance}}}^{\text{Internal~Error}}) \cdot \bf{a}^{\ell - 1})
\end{align}

Where $\eta$ is the learning rate. The convergence of this algorithm
depends on the product of the relevance signal and the activation of
the neuron, which is inherently stable from a mathematical viewpoint.
However, the experimental stability and convergence of the learning
depend on various factors such as the network's topology, weight
initialisation, experimental set-up, amplitude of inputs, nature of
the error signal, and more. Therefore, the algorithm's convergence is
best examined by observing weight changes, which are presented in the
results section.

A mathematical comparison between the two learning paradigms reveals
that they differ only in the calculation of the internal error. The
internal error of the neuron is formed by the product of the sign and
the relevance signal, analogous to that of the \ac{GDM}. In both
algorithms, the sign of the internal error is derived from the full
propagation of the control error. The distinction lies in the
calculation of the magnitude of the internal error. In the case of
\ac{GDM}, the magnitude is derived from a full propagation, which
renders the algorithm susceptible to the vanishing or exploding
gradient problem. In contrast, for \ac{SaR}, the magnitude is derived
from a single-layer propagation of the control error, making the
algorithm immune to the aforementioned problem.

From a computational standpoint, the SaR algorithm enables full
parallelisation since the calculation of the internal error in each
layer does not rely on its computation in the deeper layer.
Additionally, the sign signal can be calculated using integer
operations, which require less memory than float or double types.

\begin{figure}[t]
\centering
\includegraphics[width=0.5\textwidth]{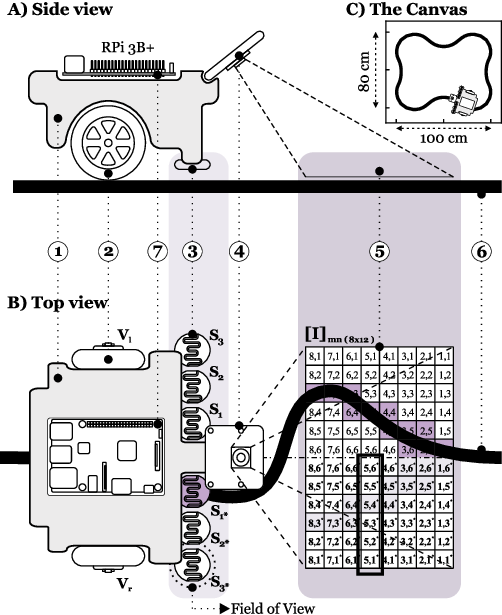}
\caption{\textit{Schematics of the navigational robot (not to scale):
A) Side view B) Top view; showing the chassis
\textcircled{\small{$1$}}, wheels \textcircled{\small{$2$}}, an
array of six \acfp{LDR} \textcircled{\small{$3$}}, camera
\textcircled{\small{$4$}}, predictive matrix
\textcircled{\small{$5$}}, path \textcircled{\small{$6$}}, and a
\acf{RPi} \textcircled{\small{$7$}} C) The canvas and the path
followed by the robot.}}\label{fig:fig4SAR_robot}
\end{figure}

\section{Implementation of \ac{SaR} learning on a navigational robot}

The learning algorithm presented in this study is implemented on a
path-following robot, referred to as the organism, with the objective
of navigating a canvas while maintaining a symmetrical position on a
path printed on the canvas, also known as the environment. Refer to
Figure~\ref{fig:fig4SAR_robot}C for an illustration.

\subsection{The robotic platform}

Schematic drawings of the robot are depicted in
Figure~\ref{fig:fig4SAR_robot}A and B (not to scale). The robot's
chassis houses a battery and contains the wiring for its components.
The robot is comprised of two wheels, an array of light sensors, and a
camera that provides a vision of the path ahead in the form of a
matrix. The \ac{SaR} algorithm is hosted on a \acf{RPi} with a remote
connection, which serves as the processing center. In the following
subsections, both reflex and predictive sensory inputs, as well as the
motor command, are described, and the components of the robot are
discussed in detail.

\subsection{Reflex sensory inputs}

The light array comprises of 6 \acfp{LDR} symmetrically positioned
underneath the chassis in close proximity to the canvas, as shown in
Figure~\ref{fig:fig4SAR_robot}A, labelled $S_{1,2,3}$ on the left and
$S_{1^{*},2^{*},3^{*}}$ on the right\footnote{The star sign denotes
the symmetrical positioning of a sensor with respect to its unmarked
counterpart.} side of the robot, as illustrated in
Figure~\ref{fig:fig4SAR_robot}B. Each sensor detects the reflected
light from a small portion of the canvas directly underneath it,
referred to as the \acf{FoV} of the sensor, as indicated by a dotted
circle around sensor $S_{3^{*}}$ in Figure~\ref{fig:fig4SAR_robot}B.
During navigation, these sensors convert the changes in the intensity
of the reflected light into voltage fluctuations. As the \acs{FoV} of
a sensor transitions from capturing the black path entirely to
capturing the white background, it generates a voltage potential
within the range of $(600-1500)[mV]$. The \acf{G}, denoted as $G_{i}
\text{ or } G_{i^{*}}$ depending on the sensor, is determined by
linearly mapping the voltage potentials of each sensor to the range
$[0,256)\in\mathbb{N}$, which represents the \acf{GSV} of its
corresponding \ac{FoV}. The \acl{G} of a sensor is proportional to the
presence of a black path in its \ac{FoV}, which provides an indication
of the sensor's vertical alignment with respect to the path and serves
as an indicator for the robot's relative positioning. For instance, in
Figure~\ref{fig:fig4SAR_robot}B, sensor $S_{1^{*}}$ is vertically
aligned with the path, indicating a slight overall deviation of the
robot to the left, whereas the alignment of sensor $S_{3}$ would
indicate a significant overall deviation to the right.

In technical terms, the robot's deviation from the path is quantified
by taking a weighted sum of the differences in the \aclp{G} of sensor
pairs. Therefore, the experimental value of the \acf{cE}, which was
previously defined in Equation~\ref{eq:cError}, can be computed as
follows:

\begin{align}\label{eq:clE}
\acs{cE} =
\Sigma_{i=1}^{3}K_{i}(\ac{G}_{i}-\ac{G}_{i^{*}}) \quad
\footnotesize{[\text{\acs{GSV}}]}
\end{align}

Here, $K_{i}$ is a weighting factor that increases linearly with $i$
to reflect the degree of deviation. This means that the farther the
active sensor is from the centerline, the greater the spike in
\ac{cE}, indicating a greater deviation. It is worth noting that, by
design, there is no ground sensor at the center of the robot to detect
the path when it is directly underneath the robot. The learning
algorithm does not rely on explicit feedback when the robot
successfully aligns with the path. The sole driver of the learning
process is the error signal generated when the robot deviates from the
path. Note that this error signal contains information about the
direction of the solution, as its sign indicates whether the robot has
veered to the left or the right of the track. This is similar to the
approach presented in \citet{10.1162/neco.2007.19.10.2694}, where the
error signal was utilised for training a single neuron, rather than a
deep network.

\subsection{Predictive sensory inputs}

The camera provides information about the path in the near distance,
which enables the robot to anticipate changes in the path and steer
accordingly. The camera captures an image with dimensions of $1280
\text{ by } 720$ pixels, which is segmented into regions as shown in
Figure~\ref{fig:fig4SAR_robot}B. Each square region is assigned the
average \acp{GSV} of the pixels it contains, generating the sensory
input from the camera in the form of an $8 \text{ by } 12$ matrix
$[I]_{mn}$. Similar to the light array sensors, the anticipated
deviation in the near distance can be inferred from the difference of
symmetrical entries in this matrix:

\begin{align} &C_{ij} = I_{ij} - I_{ij^{*}} \\ &
\text{\footnotesize{where $\quad 1 \leq i \leq m, \quad 1 \leq j
\leq \floor{\frac{n}{2}} \quad$ \text{and} $\quad j^{*} + j = n + 1$}}
\nonumber
\end{align}

Here, the $C_{ij}$ values represent the camera signals that form an $8
\text{ by } 6$ matrix. A non-zero $C_{ij}$ value indicates an upcoming
turn, where the value of $j$ indicates the sharpness of the turn, the
value of $i$ indicates the distance of the turn from the current
position, and the sign of $C_{ij}$ indicates a right or left turn.
Each difference signal is delayed using a \acf{FA} of 5 \ac{FIR}
filters, denoted as $F_{h}$, to achieve an optimal correlation with
the \acl{cE} signal for learning \citep{daryanavard2020closed}:

\begin{align}\label{eq:pk}
&p_{k} = F_{h} * C_{ij} \\ \nonumber
&\text{\footnotesize{where $\quad 1 \leq h \leq 5, \quad 1 \leq i
\leq m, \quad 1 \leq j \leq \floor{\frac{n}{2}}$}}
\end{align}

This leads to a sequence of 240 predictor signals $P_{k}$ which are
fed into the \acf{SaR} network to predict and generate anticipatory
actions.

\subsection{The \acf{MC}}

The robot's navigation is facilitated by adjusting the speed of its
right and left wheels, $V_{R}$ and $V_{L}$, which would otherwise move
forward at a fixed speed of $V_{0} =
5\textit{\footnotesize{$[\frac{cm}{s}]$}}$. A \acf{MC} is sent to the
wheels to modify their velocities as follows:

\begin{align}\label{eq:vrvl}
\begin{cases} V_{R}=V_{0}+\ac{MC}, &
\text{for right wheel}.\\ V_{L}=V_{0}-\ac{MC}, & \text{for left
wheel}.
\end{cases}
\end{align}

The \ac{MC} is generated through the joint operation of both the
reflex and predictive mechanisms of the robot. As discussed in above
the \ac{MC} is the sum of reflex and predictive actions:

\begin{align}\label{eq:mc}
\acs{MC} = A_{R} + A_{P}
\end{align}

where $A_{R}$ is proportional to \ac{cE} and $A_{P}$ is a weighted sum
of the activations in the output layer, as indicated by
Equation~\ref{eq:ap}.

\begin{align}\label{eq:motorC}
\begin{cases}
A_{R} \propto \ac{cE}, & \text{reflex action}.\\
A_{P} = f(\bf{a}^{L}) = [M] \cdot \bf{a}^{L}, & \text{predictive action}.
\end{cases}
\end{align}

The weighting matrix $[M]$ allows for sharp, moderate, or slow
steering of the robot, depending on the active neuron and its weight
factor.

At the initial stages of a trial, $A_{R}$ is the main contributor to
the \acl{MC}. As learning progresses, $A_{P}$ delivers a more adequate
contribution. Upon successful learning, where the \ac{SaR} network has
generated the forward model of the reflex, $A_{R}$ is kept at zero at
all times, and $A_{P}$ alone controls the \acl{MC}.

\subsection{The architecture of the \acf{SaR} network}

This application uses a fully-connected feed-forward neural network
with multiple layers. The network is initialised with random weights,
but it does not use any bias terms (see Eq~\ref{eq:frwdProp1} and
Eq~\ref{eq:frwdProp2}). This is because the network is designed to
receive predictive inputs and generate motor commands, both of which
are DC-free difference signals. Additionally, the control error used
to train the network is also DC-free by definition. As shown in
Equation~\ref{eq:pk}, the camera captures 240 inputs, and thus, the
input layer of the network is initialised with 240 neurons. The
network has an encoder topology, with 10 hidden layers that linearly
decrease in the number of neurons: $\{\overbrace{240}^{\text{Input}},
\overbrace{13, 12, ..., 5, 4}^{\text{Hidden}},
\overbrace{3}^{\text{Output}}\}$. The output layer consists of three
neurons, and their weighted sum of activations, using $[M]=[1,3,5]$,
produces the predictive action, as shown in Equation~\ref{eq:motorC}.

\section{Results}

\begin{figure}[t]
\centering
\includegraphics[width=0.5\textwidth]{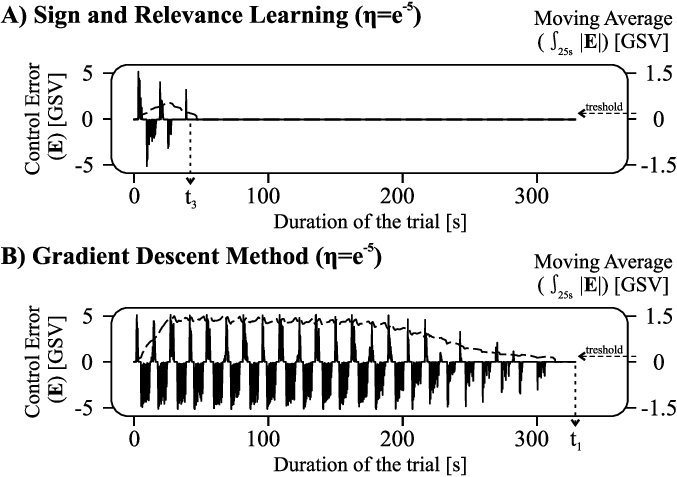}
\caption{\textit{The \acf{cE} signal and its $25[s]$ moving average
during learning trials with \ac{SaR} learning (A), and conventional
\ac{GDM} (B), using a learning rate of $\eta = e^{-5}$.}}
\label{fig:fig5SAR_result5e}
\end{figure}

In this section, we compare the conventional \ac{GDM} technique with
\ac{SaR} learning.

Figure~\ref{fig:fig5SAR_result5e} shows a set of trials with a
learning rate of $\eta=e^{-5}$. Panels A and B display the signal of
the \acf{cE} (solid traces) and its absolute moving average over 25
seconds, defined as $\bar E=\int_{t}^{t-25}|\ac{cE}(t)|$ (dashed
traces), for the two learning modalities mentioned above. Success
condition is defined as a state where $\bar E$ falls below a value of
$0.1 [\ac{GSV}]$ (left-hand-side axes), which is evaluated 12 seconds
after the trial has commenced, allowing the signal to accumulate.

In panel B, during a trial with \ac{GDM}, the error signal is
persistent for approximately 200 seconds before gradually converging
and reaching the success state at time $t_{1}=333[s]$. This trial sets
a benchmark for evaluating the \ac{SaR} paradigm. Panel A shows the
results for a trial with \ac{SaR} learning, where the top-down pass of
the sign of the error, and the magnitude of the error propagated one
layer deep, join to drive the weight changes. It can be observed that
the learning is significantly improved, as the error signal
immediately begins to converge and successful learning is achieved at
time $t_{3}=42[s]$.

\begin{figure}[t]
\centering
\includegraphics[width=0.5\textwidth]{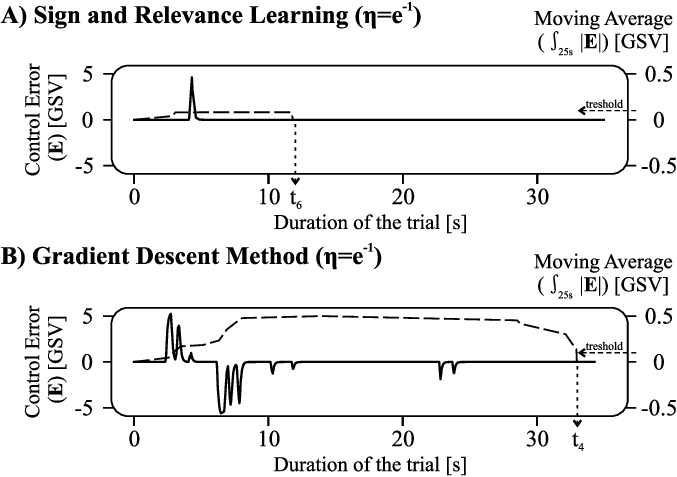}
\caption{\textit{The \acf{cE} signal and its $25[s]$ moving average
during learning trials with \ac{SaR} learning (A), and conventional
\ac{GDM} (B), using a learning rate of $\eta = e^{-1}$.}}
\label{fig:fig6SAR_result1e}
\end{figure}

Figure~\ref{fig:fig6SAR_result1e} shows another set of trials with a
faster learning rate of $\eta=e^{-1}$. During a trial with \ac{GDM} in
panel B, the error signal spikes over a period of 25 seconds before
fully converging at $t_{4}=32[s]$. Panel A shows that
\textit{one-shot} learning is accomplished during a single trial using
\ac{SaR}. The error signal spikes once at $t = 5[s]$, and the success
state is attained at $t = 12[s]$. It is evident that in this trial,
the moving average stays below the success threshold. Therefore, the
total integral of the error provides a more reliable comparative
factor, which will be depicted in Figure~\ref{fig:fig11SAR_box2}.

\begin{figure}[b]
\centering
\includegraphics[width=0.49\textwidth]{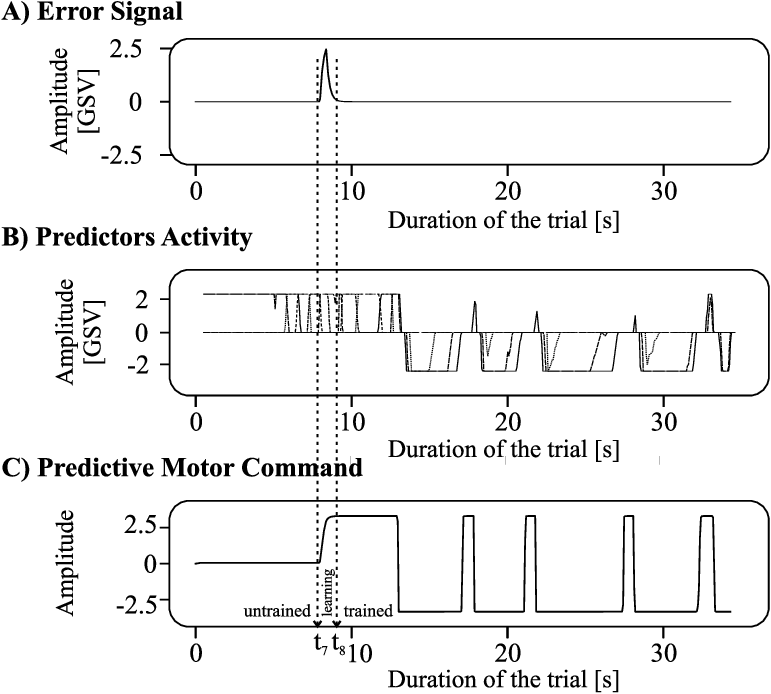}
\caption{\textit{The \acf{cE} signal (A), the activity of six selected
predictive signals (B), and the output of the neural network (C),
during a learning trial with \ac{SaR} learning using a learning rate
of $\eta = e^{-1}$.}} \label{fig:fig7SAR_predictors}
\end{figure}

Figure~\ref{fig:fig7SAR_predictors} provides additional information
about another one-shot learning trial with \ac{SaR}. Panels A, B, and
C display the \acl{cE} signal, the activity of six selected predictive
signals (marked with a rectangle in Figure~\ref{fig:fig4SAR_robot}B),
and the predictive action $A_{P}$, respectively. At time $t_{7} =
8[s]$, the robot encounters the line. Prior to this moment, some
predictive signals exhibit activity, but the network does not produce
any steering signal ($A_{P} = 0$)\footnote{This is due to the robot's
initial symmetrical position on the path.}. When the line is
encountered, the \acl{cE} experiences a spike that triggers a reflex
reaction, returning the robot to the path and training the neural
network to generate appropriate steering signals. Consequently, the
error signal returns to zero at $t_{8} = 9[s]$. Panel C shows a
gradual increase in the network output from $t_{7}$ to $t_{8}$,
indicating the learning driven by the non-zero error signal during
this interval. Once the learning is complete, the predictive signals
provide the network with clues about the upcoming path, enabling it to
generate appropriate predictive action. Consequently, the error signal
remains at zero for the rest of the trial.

Naturally, the concept of one-shot learning in the context of deep
learning raises the issue of weight stability. In this study, the
Euclidean distance between weights in each layer is utilised as an
indicator of weight convergence stability. This distance is calculated
as the multidimensional distance between the weight matrix $\bm{\omega}$
at time $t^{'}$ and its initialisation matrix at time $t_{0}$:

\begin{align}\label{eq:EDweightC}
Ed(t^{'}) = \text{Euclidean}(\omega\Big|_{t^{'}},
\omega\Big|_{init}) \\ = \sqrt[2]{\Sigma_{i,j =
0}^{I,J}(\omega_{ij}^{\ell}\Big|_{t^{'}} -
\omega_{ij}^{\ell}\Big|_{t_{0}})^{2}} \nonumber
\end{align}

This parameter is calculated within individual layers of the network
where $\ell$ is constant.

\begin{figure}[t]
\centering
\includegraphics[width=0.5\textwidth]{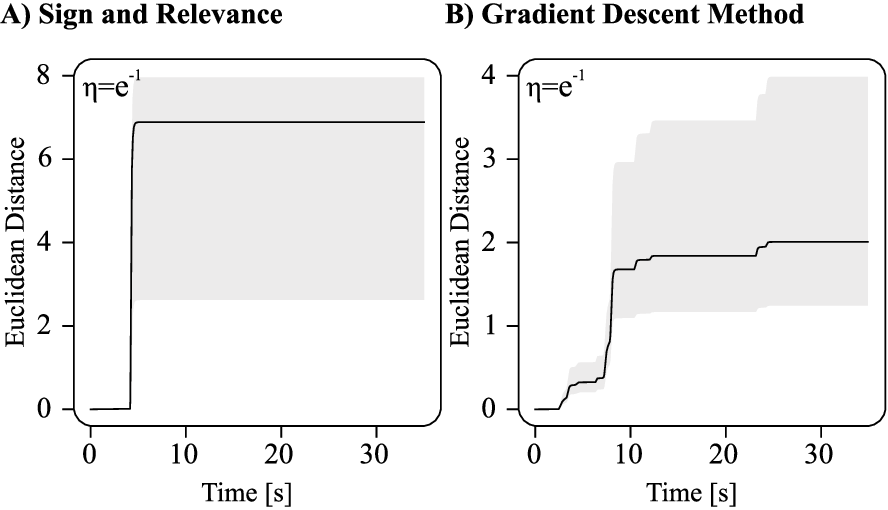}
\caption{\textit{The Euclidean distance of weights in the first layer
(black traces) and deeper layers (grey shadow) during learning trials
with \ac{SaR} learning (A), and conventional \ac{GDM} (B), using a
learning rate of $\eta = e^{-1}$.}} \label{fig:fig8SAR_weights}
\end{figure}

Figure~\ref{fig:fig8SAR_weights} depicts the Euclidean distance of
weights across the two trials previously illustrated in
Figure~\ref{fig:fig5SAR_result5e}. Panel B reveals that in a \ac{GDM}
trial, the final Euclidean distances of weights in deeper layers fall
within a range of $1.2$ to $4$ (the gray area). However, changes in
the first layer, which perceives the sensory consequences of motor
actions, bear greater importance \citep{porr2020forward}. Therefore,
the first layer's Euclidean distance is presented separately as the
black trace, with a final distance of approximately $2$ for this
trial. In Panel A, we observe this outcome for a \ac{SaR} trial where
the Euclidean distance is moderately greater than that of the
\ac{GDM}, with final distances of $3$ to $8$ for deeper layers and
approximately $7$ for the first layer. Although the Euclidean distance
has almost doubled, it has done so while improving the speed and
performance of the \ac{SaR} learner.

\begin{figure}[t]
\centering
\includegraphics[width=0.5\textwidth]{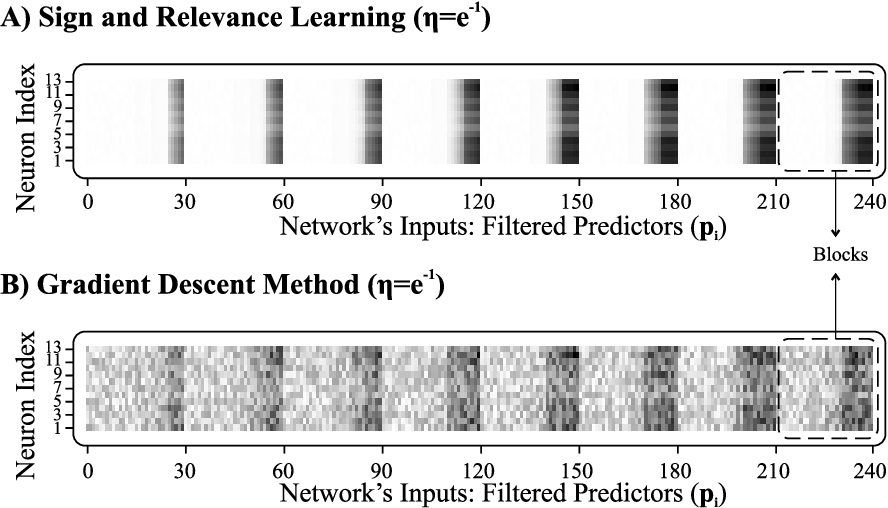}
\caption{\textit{Final weight distributions in the first hidden layer
during learning trials with \ac{SaR} learning (A), and conventional
\ac{GDM} (B), using a learning rate of $\eta = e^{-1}$.}}
\label{fig:fig9SAR_grey}
\end{figure}

\begin{figure}[b]
\centering
\includegraphics[width=0.5\textwidth]{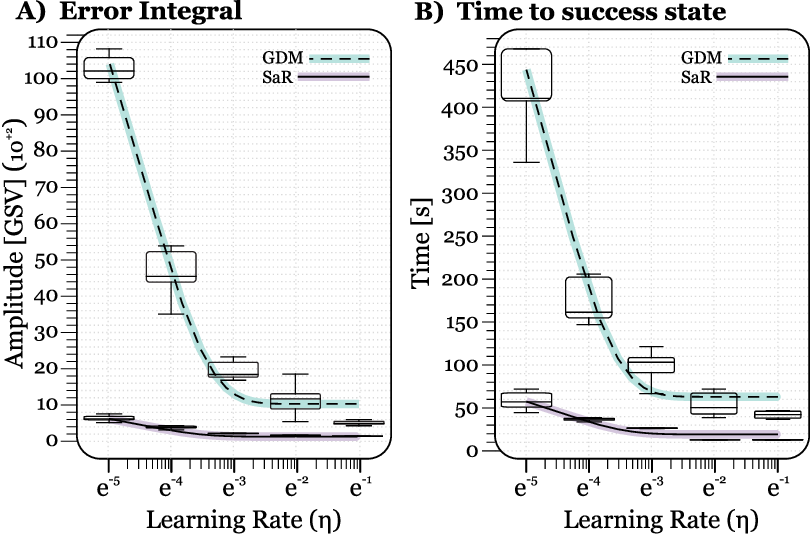}
\caption{\textit{Reproducibility of results with different learning
rates, showing the total error integral (A), and time taken to reach
the success state (B), for \ac{SaR} and \ac{GDM} learning trials.}}
\label{fig:fig10SAR_box3}
\end{figure}

\begin{figure}[b]
\centering
\includegraphics[width=0.5\textwidth]{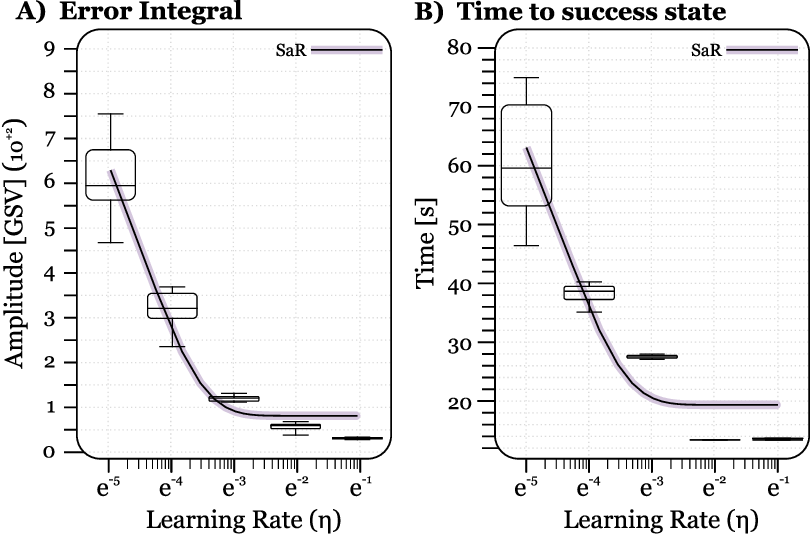}
\caption{\textit{Reproducibility of results with different learning
rates, showing the total error integral (A), and time taken to reach
the success state (B), for \ac{SaR} learning trials.}}
\label{fig:fig11SAR_box2}
\end{figure}

\begin{figure}[t]
\centering
\includegraphics[width=0.5\textwidth]{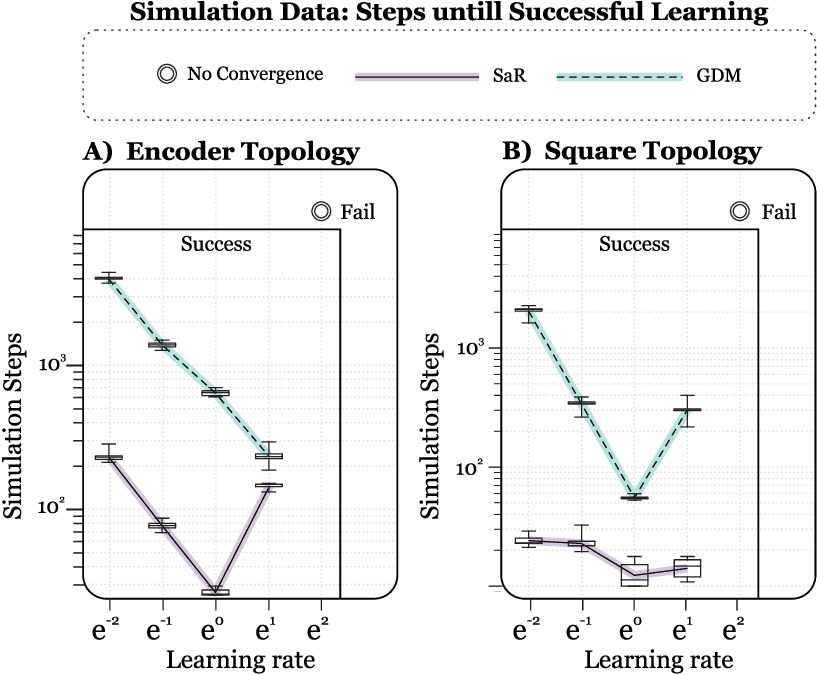}
\caption{\textit{Simulation results with higher learning rates include
the time taken to meet the success condition. Runs exceeding 10,000
steps are considered failures, indicating that the network did not
converge. (A) Results with a deep network with an encoder topology.
(B) Similar results with a square topology.}}
\label{fig:fig15SAR_eta_boxplot}
\end{figure}

As mentioned previously, the first hidden layer holds greater
significance as it is where the organism learns to assign importance
to the sensory inputs it receives from the environment. In
Figure~\ref{fig:fig9SAR_grey}, the final weight distributions in the
first hidden layer for the trials previously discussed in
Figures~\ref{fig:fig5SAR_result5e} and~\ref{fig:fig8SAR_weights} are
displayed. This layer comprises 240 predictive input signals and 13
neurons, resulting in a weight matrix of size $240$ by $13$. The
values in this matrix are normalised within the range of $[0, 1]$ and
are represented by a grey scale, with white corresponding to $0$ and
black to $1$. This creates an image of the weight distribution.

Common observations among the two learning paradigms are that 1) the
eight rows of predictors (as shown in Figure~\ref{fig:fig4SAR_robot})
are appropriately classified into recognisable blocks. 2) Each block
contains a gradient where the outermost columns of predictors (as seen
in Figure~\ref{fig:fig4SAR_robot}) are assigned a higher value,
resulting in sharper steering. 3) This gradient is more pronounced for
rows of predictors closer to the robot, particularly the rightmost
blocks. However, a comparison of the trials reveals a more confident
and well-defined gradient for \ac{SaR} learning in panel A, as opposed
to the \ac{GDM} in panel B.

\begin{figure}[b]
\centering
\includegraphics[width=0.5\textwidth]{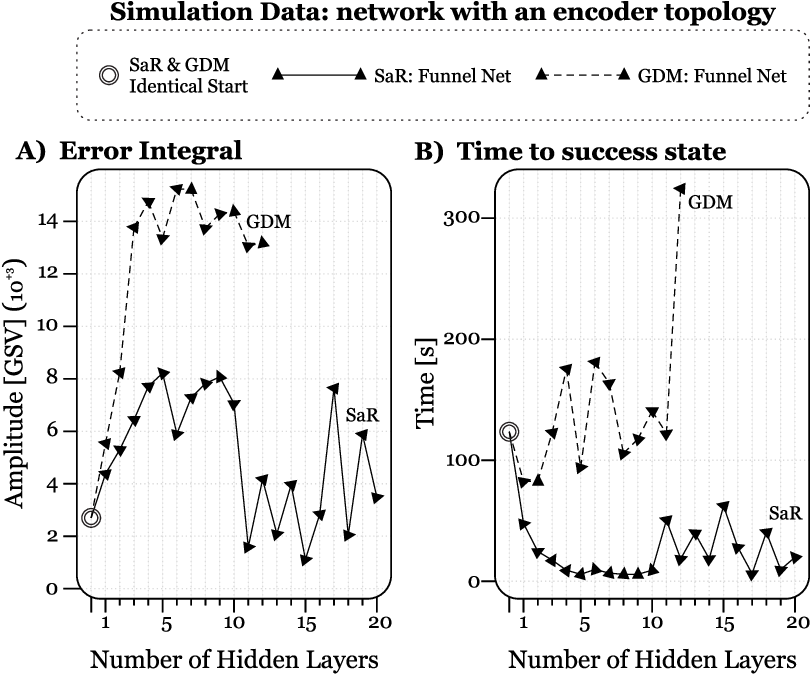}
\caption{\textit{The effect of the depth of the network, with an
encoder topology, on the total error integral (A), and the time
taken to reach the success state (B).}}
\label{fig:fig12SAR_netTriangle}
\end{figure}

\begin{figure}[b]
\centering
\includegraphics[width=0.5\textwidth]{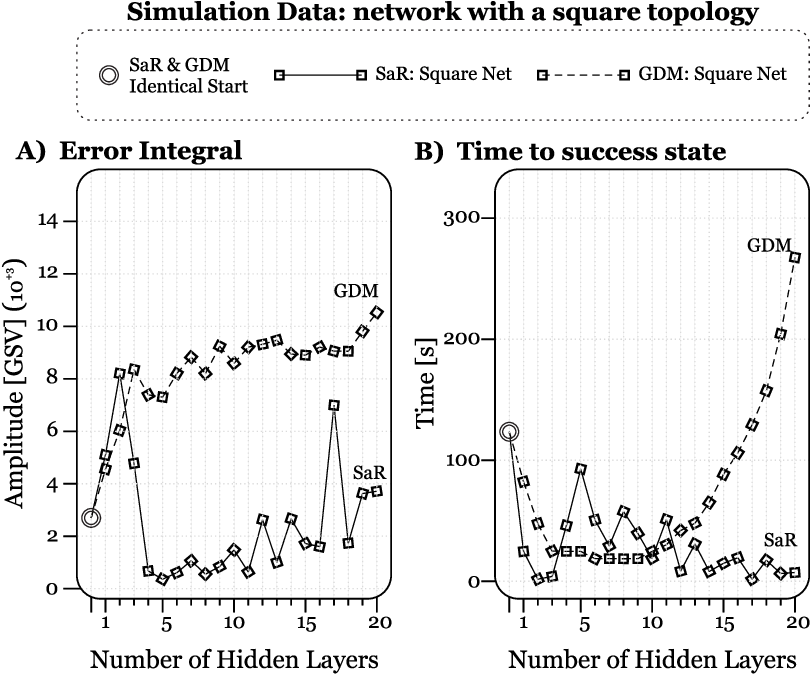}
\caption{\textit{The effect of the depth of the network, with a
square topology, on the total error integral (A) and the time
taken to reach the success state (B).}}
\label{fig:fig13SAR_netSquare}
\end{figure}

The trials depicted in Figure~\ref{fig:fig5SAR_result5e} and
Figure~\ref{fig:fig6SAR_result1e} were carried out with learning rates
of $\eta = \{e^{-5},e^{-4},e^{-3},e^{-2},e^{-1}\}$, with each case
repeated 10 times to demonstrate the reproducibility of the results.
Figures~\ref{fig:fig10SAR_box3} A and B display the total error
integral and the time taken to reach the success condition for trials
conducted with \ac{SaR} and \ac{GDM}. It can be concluded that
\ac{SaR} consistently provides faster learning with smaller error
accumulation, regardless of the learning rate. The comparison between
the two algorithms in Figure~\ref{fig:fig5SAR_result5e}
and~\ref{fig:fig6SAR_result1e} is justified by the observation that
both algorithms perform best with higher learning rates and their
performance declines with smaller learning rates. Therefore,
Figure~\ref{fig:fig6SAR_result1e} compares the two algorithms at their
best performance, while Figure~\ref{fig:fig5SAR_result5e} compares
them at their worst. Furthermore, in Figures~\ref{fig:fig10SAR_box3},
it can be observed that at lower learning rates, the advantage of
\ac{SaR} is more pronounced. The data set for each paradigm were
fitted with an exponential function of the form $y = a \cdot e^{-b
\eta} + c $, which are shown superimposed on the data points.
\footnote{The coefficients a, b, and c, for error integrals are found
to be: $GDM=(16e3,82,1e3)$, and $SaR=(1e3,91,78)$, and for success
time:
$GDM=(737,97,60)$, and $SaR=(79,83,18)$.} Both algorithms exhibit an
exponential decline with decreasing learning rates, but this decline
is more rapid for \ac{GDM} than for \ac{SaR}. For a closer inspection,
Figure~\ref{fig:fig11SAR_box2} shows a magnified plot of the \ac{SaR}
data.

For a more equitable comparison between \acs{SaR} and \acs{GDM}, we
conducted additional simulation experiments using higher learning
rates to identify the optimal settings for each of these paradigms.
Both paradigms failed to converge under the specified success criteria
when using a learning rate of $e^{2}$. For completeness, we considered
learning rates in the set $\eta = \{e^{-2}, e^{-1}, e^{0}, e^{1},
e^{2}\}$, and repeating the experiments $ 10 $ times. The results of
these experiments are presented in
Figure~\ref{fig:fig15SAR_eta_boxplot}. In Panel A, the results are
shown for a deep neural network with an encoder topology. \acs{GDM}
performs best at a learning rate of $ e^{1} $, reaching an average of
$ 288 $ steps before convergence, while \acs{SaR} performs optimally
at $ e^{0} $, requiring an average of $ 37 $ steps for convergence.
Panel B illustrates a similar outcome for a network with a square
topology. Here, both paradigms achieve their best performance at a
learning rate of $ e^{0} $. \acs{GDM} converges in an average of $ 64
$ steps, while \acs{SaR} achieves success in an average of $ 15 $
steps.

Figure~\ref{fig:fig12SAR_netTriangle} illustrates the effect of the
number of layers in a network with a typical triangle-shaped encoder
topology. The solid line represents \ac{SaR} and the dashed line
represents \ac{GDM}. The number of hidden layers increases from 0 to
20, with the number of neurons increasing linearly for each added
layer: $\{\overbrace{240}^{\text{Input}}, \overbrace{23, 22, ..., 5,
4}^{\text{Hidden}}, \overbrace{3}^{\text{Output}}\}$. Zero hidden
layers indicate that the network is initialised with one input layer
and one output layer, where inputs are directly fed into the output
neurons. Mathematically speaking, there is no difference between the
two algorithms when no hidden layers are present. This has been
experimentally demonstrated, as both algorithms produce identical
results. Interestingly, the \ac{SaR} algorithm is less affected by
variations in the number of layers, as expected from its single-layer
propagation. On the other hand, the \ac{GDM} algorithm shows a
decrease in performance as the number of layers increases. It is worth
noting that there are no data points for \ac{GDM} with more than 12
hidden layers since the success condition was not met (convergence was
not achieved), which is due to the vanishing gradient problem. It is
evident that the \ac{SaR} network continues to perform well with as
many as 20 hidden layers without significant changes in the results
shown.

In a triangle-shaped encoder network, the total number of neurons
increases drastically with each added layer. To make a more accurate
comparison between the two paradigms, another set of simulations was
conducted using a square-shaped network where the number of neurons
was fixed at 10 per layer for each added layer:
$\{\overbrace{240}^{\text{Input}}, \overbrace{10, 10, ..., 10,
10}^{\text{Hidden}}, \overbrace{3}^{\text{Output}}\}$.
Figure~\ref{fig:fig13SAR_netSquare} illustrates the results of these
experiments. As the number of hidden layers increases, both algorithms
show some improvement. However, as the number of layers continues to
grow, the \ac{SaR} algorithm is less affected by the depth of the
network, whereas the \ac{GDM} algorithm progressively declines in
performance. For \ac{GDM}, the time taken to achieve success grows
exponentially with an increasing number of hidden layers, and the
error integral shows a linear growth. This indicates that \ac{GDM}
suffers from the vanishing gradient problem in deep networks, unlike
its \ac{SaR} counterpart.

A comparison of the two topologies reveals that both algorithms
perform better with a square-shaped network where the number of
neurons is the same across all layers. This is expected because any
increase in the number of neurons in triangle-shaped encoder networks
directly affects the weighted sums of inputs and the internal errors
in the forward and backward passes, respectively. As a result, the
propagation is more susceptible to unstable changes, which can
negatively impact performance.

\section{Discussion}\label{sec:discussion}

\begin{figure}[t]
\centering \includegraphics[width=0.48\textwidth]{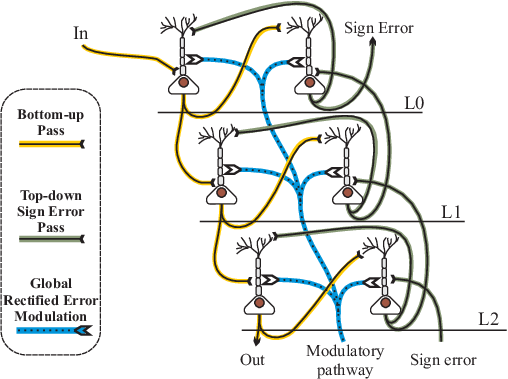}
\caption{\textit{Proposal of a neurophysiologically realistic model of
\ac{SaR} learning. The figure shows three network layers (L0, L1, and
L2), with signal processing occurring in three pathways: the
``bottom-up'' pathway, which transmits a signal from ``In'' to
``Out''; the ``top-down'' pathway, which transmits the \textit{Sign}
error; and the ``modulatory'' pathway, which provides a global signal
to all neurons. The bottom-up and top-down pathways transmit signals
via synapses close to the respective somas, while the reciprocal
connections between neurons within a layer connect to the dendrites,
influencing plasticity.}} \label{fig:fig14SAR_model}
\end{figure}

In this paper, we introduce a learning algorithm that decomposes the
error signal into a global and sign component. The sign component is
passed in a top-down manner through the network, while its rectified
value is transmitted globally to all layers. Unlike classical
back-propagation, this approach is significantly faster in a
closed-loop learning task and more closely aligns with neurophysiology
by incorporating both neuromodulators and calcium-driven \acf{ltd} or
\acf{ltp}. We demonstrate successful learning in a simple
line-following task, and this approach can be applied to any task
where a fixed closed-loop ``reflex'' controller generates an error
signal. Additionally, this approach potentially enables multi-modal
processing, as any input can be used to learn a forward model of the
reflex.

Deep learning has gained widespread popularity over the last decade.
As it utilises neural networks, it is a promising candidate to explain
how the brain itself learns \citep{marblestone2016toward}. For
instance, \citet{Lillicrap2016} has mapped deep learning onto the
cortex, but has not considered global neuromodulation such as
serotonin or dopamine.

On the other hand, traditional biologically realistic reinforcement
learning models \citep{Suri2001, Woergoetter2005, Prescott2006robot}
employ the reward prediction error, which bears a strong resemblance
to the dopaminergic signal in the striatum \citep{Schultz97}. These
models are certainly closer to biology, but they suffer from the
problem that any global error signal poses for deep structures. That
is, the different layers change similarly, and hence, deep structures
do not add much to their performance.

However, it remains unclear whether reward-related neuromodulators
convey error signals, as has been the dominant paradigm over the last
two decades \citep{Schultz97}. Serotonin, for example, appears to
encode both reward and punishment expectation
\citep{Li2016,Crockett2009,10.7554/eLife.06346}. We have
mathematically subsumed this into a ``modulus'', but in the
neurophysiological context, it may be more appropriate to refer to it
as a ``relevance signal'' because it switches on plasticity, as
suggested by \citet{Porr2007}. Similarly, the negative response of
dopamine neurons to a negative reward expectation has been deemed
unreliable by \citet{schultz04} themselves, due to its low baseline
firing rate of approximately 1 Hz, which leads to a very low
signal-to-noise ratio. A different interpretation for both the
serotonin and dopamine signals is that of a relevance signal
\citep{Porr2007}, ramping up or enabling plasticity
\citep{Lovinger2010, Iigaya2018}, while local plasticity learning
rules determine if synaptic weights undergo \acf{ltp} or \acf{ltd}
\citep{Castellani01,Inglebert2020}.

Having established global neuromodulation and local plasticity, we can
now discuss how such processing can be implemented in a biologically
realistic fashion. Figure~\ref{fig:fig14SAR_model} depicts the
suggested circuit, inspired by the work of
\citet{larkum2013cellular,rolls2016reward}, and with added
neuromodulatory innervation \citep{Lovinger2010,Iigaya2018}. This
circuit consists of two distinct pathways. The bottom-up pathway
conveys, for example, sensor signals to deeper brain structures or
directly to motor outputs. A single bottom-up path is depicted from
the input, labelled ``In'', to the output, labelled ``Out'', via three
synapses connecting the three neurons in layer zero (L0) to layer two
(L2). Having described the bottom-up path, we can now turn to the
top-down path, which controls the \textsl{sign} of learning (as
defined in Eq.~\ref{eq:signcalc}) and ultimately determines which
neurons undergo \acf{ltp} or \acf{ltd}. The bottom-up pathway
transmits the sign of the error signal from layer two back to layer
zero. This pathway also consists of three synapses. Additionally,
there is global neuromodulation $E$ that controls the plasticity of
all neurons (as defined in Eq.~\ref{eq:gamma}). To link this to
neurophysiology, we use the well-established mechanism for driving
standard neuronal plasticity, which is the concentration of
\textsl{postsynaptic calcium}. To justify our binary switch between
\acf{ltp} and \acf{ltd} (as defined in Eq.~\ref{eq:signcalc}), we
follow the reasoning of \citet{Inglebert2020}: only a strong calcium
influx caused by both somatic burst spiking and dendritic calcium
spikes will result in \acf{ltp}, while less in to \acf{ltd}
\citep{Tamosiunaite2007}. Conventional backpropagation dictates that
the plasticity changes for each synapse depend on the precise
magnitude of the synaptic changes of deeper neurons, necessitating
symmetrical weights in the both pathways. However, the network-wide
binary propagation of the internal error (Equation~\ref{eq:signcalc})
relaxes this requirement, as neurons only need to determine whether
the deeper synapse underwent \acf{ltd} or \acf{ltp}.

Projections from the distal parts to the dendrites alone are
insufficient to induce spiking in neurons. However, if they coincide
with somatic inputs, they can cause long-lasting bursting that leads
to the induction of \ac{ltp} due to a large influx of calcium
\citep{larkum2013cellular}. Conversely, insufficient activation of
dendritic trees can initiate \ac{ltd} due to a smaller calcium influx
from single spikes \citep{Inglebert2020,Shouval02}. Now, let us
consider the reciprocal connections between neurons, such as those
observed in L1, as depicted in Figure~\ref{fig:fig14SAR_model}. If the
neuron in the top-down pathway exhibits strong spiking activity due to
high synaptic weight from L2, it will also drive the neuron in the
bottom-up pathway to spike, causing a calcium influx. A strong influx
will trigger \ac{ltp} in the bottom-up pathway, while a small influx
will result in \ac{ltd}. Increased synaptic weights in the bottom-up
pathway will cause strong activity, which will, in turn, drive the
top-down pathway and result in more calcium influx. Therefore, the
\textsl{sign} of the weight development between reciprocal neurons in
both the top-down and bottom-up pathways is mirrored because of their
reciprocal connections and their ability to boost or deprive each
other's calcium concentrations. The neuromodulator $E$ controls the
learning rate by regulating the amount of either \acf{ltp} or
\acf{ltd}. Further research is required to investigate this model in
depth using more detailed biophysical modelling, which could have
positive implications for models of mental illness
\citep{rolls2016reward}.

We predict that both global neuromodulation and local calcium-driven
plasticity are necessary for successful behavioural adaptation in
instrumental or Pavlovian learning experiments. In particular,
successful learning in the cortex requires a combination of local
calcium-driven plasticity and neuromodulation. Blocking the NMDA
receptor disrupts calcium-driven plasticity, and we predict that
cortical reward- or punishment-based learning would also be disrupted
as a result. This could suggest alternative drugs that target cortical
calcium/NMDA-driven plasticity rather than neuromodulators,
potentially offering new treatments for depression. If the sign of
learning (i.e., LTP or LTD) is determined in the cortical circuitry
and propagated through its network, disrupting this kind of error
propagation would still allow for learning in general via the
neuromodulator, but it would no longer be directed towards a learning
goal, such as finding a reward or avoiding an aversive stimulus.
Instead, learning would occur without a specific goal, which would
limit its usefulness.

\begin{dci}
There is no conflict of interest in this work.
\end{dci}

\begin{funding} We would like to thank Engineering and Physical
Sciences Research Council (EPSRC) division of UK Research and
Innovation (UKRI) for funding (EP/N509668/1 and EP/R513222/1) this
project.
\end{funding}

\begin{acks} We would like to acknowledge Jarez Patel for his valuable
intellectual and technical input for the making of the robotic
platform.
\end{acks}

\bibliographystyle{SageH} 
\bibliography{refs_SaR.bib}

\end{document}